\documentclass[journal]{IEEEtran}
\usepackage{graphicx}
\usepackage{amssymb,amsmath,bm}
\usepackage{textcomp}
\usepackage{amsmath}
\usepackage{graphics}
\usepackage{epsfig}
\usepackage{multirow}
\usepackage{amsmath}
\usepackage{enumitem}
\usepackage{mathrsfs}
\usepackage{bm}
\usepackage{epstopdf}
\usepackage{array}

\usepackage{float}

\ifCLASSINFOpdf
\else
\fi
\begin{document}
%
\title{Progressive Joint  Modeling in Unsupervised Single-channel Overlapped Speech Recognition}
%
%
%
%

\author{
Zhehuai~Chen,~\IEEEmembership{Student Member,~IEEE,},
Jasha~Droppo,~\IEEEmembership{Senior Member,~IEEE,},
Jinyu~Li,~\IEEEmembership{Member,~IEEE,},
Wayne~Xiong,~\IEEEmembership{Member,~IEEE,},

\thanks{
}
\thanks{
Zhehuai\,(Tom) Chen is with the Computer Science and Engineering Department, Shanghai Jiao Tong University, Shanghai 200240, China.
He conducted this work while working as an intern at Microsoft Research. (email: chenzhehuai@sjtu.edu.cn) 

Jasha Droppo, Jinyu Li and Wayne Xiong 
are
with Microsoft AI and Research, One Microsoft Way, Redmond, WA 98052,
U.S.A. (e-mail: \{jdroppo,jinyli,wayne.xiong\}@microsoft.com)


Corresponding authors are Jasha Droppo and Zhehuai Chen (email: jdroppo@microsoft.com,chenzhehuai@sjtu.edu.cn  phone: +1 425 703-7114).

}
}

\maketitle

\begin{abstract}

Unsupervised single-channel overlapped speech recognition is one of the hardest problems in automatic speech recognition (ASR). Permutation invariant training (PIT) is a state of the art model-based approach, which applies a single neural network to solve this single-input, multiple-output modeling problem.
We propose to advance the current state of the art by imposing a modular structure on the neural network, applying a progressive pretraining regimen, and improving the objective function with transfer learning and a discriminative training criterion. The modular structure splits the problem into three sub-tasks: frame-wise interpreting, utterance-level speaker tracing, and speech recognition. The pretraining regimen uses these modules to solve progressively harder tasks. Transfer learning leverages parallel clean speech to improve the training targets for the network. Our discriminative training formulation is a modification of standard formulations that also penalizes competing outputs of the system.
Experiments are conducted on the artificial overlapped Switchboard and hub5e-swb dataset. The proposed framework achieves over 30\% relative improvement of WER over both a strong jointly trained system, PIT for ASR, and a separately optimized system, PIT for speech separation with clean speech ASR model. The improvement comes from better model generalization, training efficiency and the sequence level linguistic knowledge integration.

\end{abstract}



\IEEEdisplaynontitleabstractindextext

%
\IEEEpeerreviewmaketitle


\section{Introduction}\label{Sec:}

The cocktail party problem \cite{cherry1953some,bregman1994auditory},  referring to multi-talker overlapped speech  recognition, 
is critical 
to enable automatic speech recognition (ASR) scenarios such as automatic meeting transcription, automatic
captioning for audio/video recordings, and multi-party human-machine interactions, where overlapping speech  is commonly observed and all streams need to be transcribed.
The problem is still one of the hardest problems in ASR, 
despite  encouraging progresses~\cite{wang2006computational,cooke2010monaural,du2014speech,weng2015deep}. 
%

In this paper, we address the speech recognition
problem  when multiple people speak at the same time and only
a single channel of overlapped speech is available. This is useful when only a single microphone is present, or when microphone array based algorithms fail to perfectly separate the speech.
Specifically, the paper  focuses on an unsupervised inference method, which does not need any prior knowledge of speakers. 
To obtain transcriptions of all speakers from the  overlapped speech, joint inference is conducted based on  multiple knowledge sources: frequency domain voice discrimination, temporal speaker tracing, linguistic information and speech recognition. 

Prior work in unsupervised
single-channel overlapped speech recognition 
generally  separates the problem into 
speech separation and recognition stages. 
Before the deep learning era, 
the most popular speech separation technique
is computational auditory scene analysis (CASA)~\cite{wang2006computational}. 
There are two main stages in CASA approaches: segmentation
and grouping. 
The segmentation stage decomposes mixed speech
into time-frequency segments assumed to be derived from
the corresponding speakers based on perceptual grouping
cues~\cite{wertheimer1938laws}.
The grouping stage simultaneously and sequentially
 concatenates the segments to generate independent
streams for each speaker.
Non-negative matrix factorization (NMF)~\cite{schmidt2006single} is another popular technique which aims to learn a set
of non-negative bases that can be used to estimate mixing factors
during evaluation.
%
Recently, several deep learning based techniques have been proposed but seldom concentrate on the unsupervised case, which is more applicable.
In~\cite{hershey2016deep,chen2017deep,isik2016single}, the authors propose deep clustering (DPCL), 
in which a deep network is trained to produce spectrogram embeddings that are
discriminative for partition labels given in training data.
The model is optimized so that in the neural network embedding space the time-frequency bins belonging to the same speaker are closer and those of different speakers are farther away.  
Speech segmentations are therefore implicitly encoded in the embeddings, and can be obtained by clustering algorithm.
In~\cite{wang2017gender},  
a DNN-based gender mixture detection system and three gender-dependent  speech separation systems are constructed. The latter ones directly infer the feature streams  of two speakers respectively.
For all these methods,
speech
separation and recognition are two separate components and the latter is applied to  the separated feature streams. The mismatched feature  in the speech recognition stage is one of the limitation in these methods.

In light of permutation invariant training (PIT)  proposed in speech separation~\cite{yu2017permutation} originally, the PIT-ASR model~\cite{yu2017recognizing}   is the first attempt in joint modeling of unsupervised single-channel mixed speech recognition.
Whereas the original PIT technique jointly models the voice discrimination and speaker tracing, PIT-ASR further integrates speech recognition into the neural network with a unified cross entropy (CE) criterion.
Although PIT-ASR shows promising results, it suffers from several disadvantages, which are analyzed in Section~\ref{Sec:review-si-ch-rec}.

\begin{figure*}
  \centering
    \includegraphics[width=0.8\linewidth]{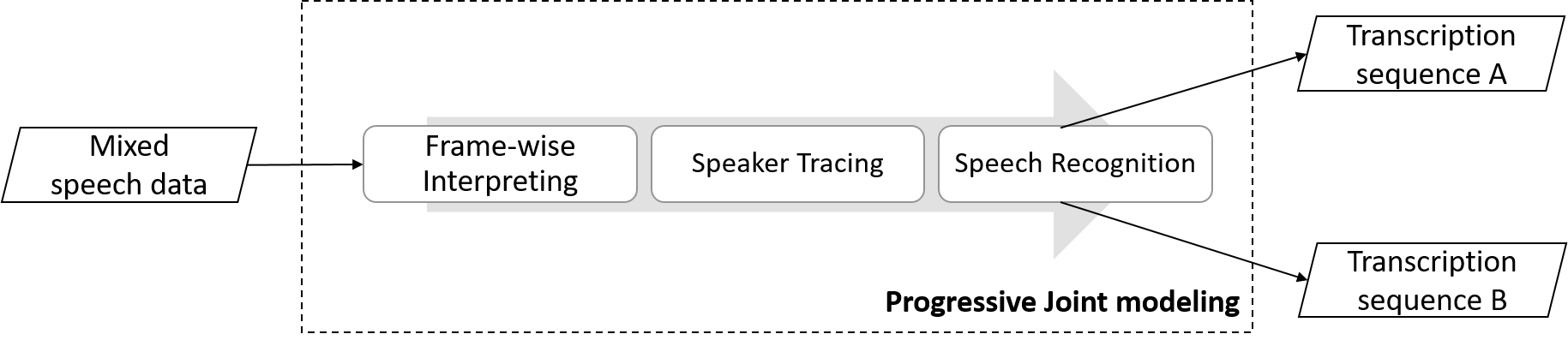}
    \caption{\it The Proposed System Framework. 
    The single monolithic structure (the dashed line box) that predicts independent targets for each speaker, proposed in~\cite{yu2017recognizing}, is improved through modularization (three solid line boxes) and pretraining.
    Self-transfer learning and multi-output sequence discriminative training  are conducted on modularly initialized layers.}
    \label{fig:sys-fr}
\end{figure*}

In this paper, progressive joint modeling is proposed to divide
the single channel overlapped speech recognition problem
into three sub-problems for initialization: frame-wise interpreting, speaker tracing and speech recognition (Figure~\ref{fig:sys-fr}). 
Each module is initialized by placing it into a series of networks that solve progressively more difficult problems.
After the initialization, modules are jointly trained with two novel strategies, namely
self-transfer learning and multi-output sequence discriminative
training. 
Transfer learning is introduced in this problem, which  leverages parallel clean speech to improve the training targets for the network. Our discriminative training formulation is a modification of standard formulations, that also penalizes competing outputs of the system. 
The proposed framework
achieves 30\% relative improvement over both a strong jointly
trained system, PIT-ASR, and a separately optimized system,
PIT for speech separation with clean speech ASR. The improvement comes from better model
generalization, training efficiency and the sequence level linguistic knowledge integration.

The rest of the paper is organized as follows. In Section~\ref{Sec:review-si-ch-rec}, the unsupervised single-channel overlapped speech recognition problem is briefly reviewed. In Section~\ref{Sec:modu-init}, the modular initialization and progressive joint training is proposed. In Section~\ref{Sec:self-transf}, the self-transfer learning is proposed and in Section~\ref{sec:seq-disc-tr}, multi-output sequence discriminative training is proposed. In Section~\ref{Sec:exp}, the experimental results are reported in artificial overlapped Switchboard corpus and Eval2000 hub5e-swb test set, followed by the conclusion in Section~\ref{Sec:conclu}.


\section{Unsupervised Single-channel Overlapped Speech Recognition}\label{Sec:review-si-ch-rec}

Unsupervised single-channel overlapped speech recognition refers to 
the speech recognition
problem when multiple unseen talkers speak at the same time and only
a single channel of overlapped speech is available. Different from supervised mode, there's not  any prior knowledge of speakers in the evaluation stage. 

In the problem, only the linearly overlapped single-channel signal 
is known, which is defined as,

\begin{equation}
\label{equ:mix-data}
\begin{split}
\mathbf{O}_{u}^{(m)}=\sum_{n=1}^N \mathbf{O}_{un}^{(r)}
\end{split}
\end{equation}
where $\mathbf{O}_{un}^{(r)}$ is the clean signal stream of speaker $n$ at utterance $u$ and $\mathbf{O}_{u}^{(m)}$ is the overlapped speech stream of utterance $u$. $N$ is the number of streams.
Single channel ASR is always formulated as a  supervised sequence labeling problem given by $P(\mathbf{L}_u|\mathbf{O}_{u})$, which is the posterior probability of transcription sequence $\mathbf{L}_u$ given the feature sequence $\mathbf{O}_{u}$ in the utterance $u$.
Nevertheless,  the multi-speaker  problem is to model the  joint distribution of $N$ streams of transcriptions given the overlapped speech signal,  $P(\mathbf{L}_{u1},...,\mathbf{L}_{uN}|\mathbf{O}_{u}^{(m)})$. 
Due to the  symmetric labels  given the
mixture signals, it is no longer  a  supervised optimization problem. 
One branch of methods assumes the inference streams are conditionally independent, and tries to assign the correct transcription stream $\mathbf{L}_{un}^{(r)}$ to the corresponding output stream $n$,
\begin{equation}
\label{equ:obj-joint}
\begin{split}
P(\mathbf{L}_{u1},...,\mathbf{L}_{uN}|\mathbf{O}_{u}^{(m)}) \approx 
\prod_{n=1}^N P(\mathbf{L}_{un}^{(r)}|\mathbf{O}_{u}^{(m)})
\end{split}
\end{equation}
Another branch of methods
assume the overlapped signal can be separated to $\mathbf{O}_{un}^{(c)}\mathop{:} \mathbf{O}_{un}^{(c)}\approx \mathbf{O}_{un}^{(r)}$. Because the original streams $\mathbf{O}_{un}^{(r)}$ are conditionally independent with each other,  the separated signal streams $\mathbf{O}_{un}^{(c)}$ are also assumed to be conditionally independent. Thus Equation~(\ref{equ:obj-joint}) can be derived to Equation~(\ref{equ:obj-sep}),
\begin{equation}
\label{equ:obj-sep}
\begin{split}
P(\mathbf{L}_{u1},...,\mathbf{L}_{uN}|\mathbf{O}_{u}^{(m)}) \approx 
\prod_{n=1}^N P(\mathbf{L}_{un}^{(r)}|\mathbf{O}_{un}^{(c)})
\end{split}
\end{equation}
However, neither assumption is precise. For the first assumption, there is no pre-determined method to  obtain the ideal label arrangements, which is called the speaker tracing problem. 
The second assumption is that the speech separation and recognition are independent processes, which introduces an artificial information bottleneck.

In \cite{yu2017permutation}, 
the reference streams are treated as an unordered set.
The PIT framework is proposed to address the speech separation problem by firstly determining the assignment of
the reference stream and inference stream 
that minimizes the  error at the
utterance level based on the forward-pass result. 
This is followed by minimizing
the error given the utterance level best assignment. 
\cite{yu2017recognizing} extends this by integrating 
speech recognition into the neural network with a  unified
 cross-entropy (CE) training criterion.
\begin{equation}
\label{equ:ce-pit}
\begin{split}
\mathcal{J}_{\text{CE-PIT}}=\sum_u \min_{s'\in \mathbf{S}} \sum_t \frac{1}{N} \sum_{n\in[1,N]} CE({l}_{utn}^{(s')},{l}_{utn}^{(r)})
\end{split}
\end{equation}
Here,  $\mathbf{S}$ is the permutation set of the reference representation and the inference representation.
${l}_{utn}^{(s')}$ is the $n$-th inference label  of permutation $s'$ at  frame $t$ in utterance $u$ and ${l}_{utn}^{(r)}$ is the corresponding transcription label obtained by clean speech forced-alignment~\cite{woodland1994large}.

The PIT-ASR criterion~\cite{yu2017recognizing} elegantly integrates speech separation, speaker tracing and speech recognition together as Figure~\ref{fig:modules}(a).  
Its joint modeling approach eliminates the artificial bottleneck between the speech separation and speech recognition tasks.
But the method suffers from several disadvantages which deteriorates  the performance:
\begin{itemize}
\item Previous PIT-ASR work attempts to solve three of the most difficult problems in speech processing with one large model. This oversized model can lead to poor generalization and slow training.
\item Speaker tracing and speech recognition  are both sequence modeling problems.
PIT-ASR  concatenates frame level CE as the criteria of the sequence level problem, which limits the modeling effect of the neural network.
\item PIT-ASR with a monolithic model is difficult to integrate with other speech separation and segmentation technologies. For instance, although 
linguistic information can be valuable for solving the speaker tracing problem~\cite{weng2015deep}, there is no natural way to incorporate it into PIT-ASR.

\end{itemize}

\section{Methods}\label{Sec:methods}

In this work, we propose three separate enhancements to improve the performance of PIT-ASR.

First, the structure and accuracy of the model is improved through modularization and pretraining. Frame-wise interpreting, speaker tracing, and speech recognition modules replace the monolithic structures used in previous work. These modules are progressively pretrained and jointly fine-tuned.

Second, we demonstrate a natural way to incorporate a form of transfer learning. Clean speech features are used to generate soft label targets which are interpolated with the reference label alignments.

Third, multi-output discriminative training is applied to the system. As with single-stream speech recognition, multi-stream discriminative training can help with model generalization. Additionally, the objective function is augmented to reduce cross-speaker word assignment errors.

\subsection{Modularization}\label{Sec:modu-init}

In the original formulation, a PIT-ASR model consists of a single monolithic structure that predicts independent targets for each speaker. We improve this by replacing the main network structure with a modular structure, shown in Figure~\ref{fig:sys-fr}.

This modular structure consists of three tasks, namely interpreting mixed acoustic data, tracing speakers across time, and predicting acoustic label sequences. First, the frame-wise module is designed to extract the local time-frequency information necessary to separate the overlapped speech into individual acoustic representations. It is entirely local and does not depend on sequence-level information. Second, the speaker tracing module accepts frame-wise acoustic representations from the frame-wise module and traces the speaker information. This process concatenates adjacent acoustic representations of the same speaker together to infer the recovered speech features of each speaker. Third, the speech recognition modules accept the sequences of recovered acoustic features from each speaker, and produce sequences of label scores suitable for use in an automatic speech recognition system. Because each speech recognition module performs the same task, it is natural to share the parameters of this module across each instance in the final model.

\begin{figure}
  \centering
    \includegraphics[width=\linewidth]{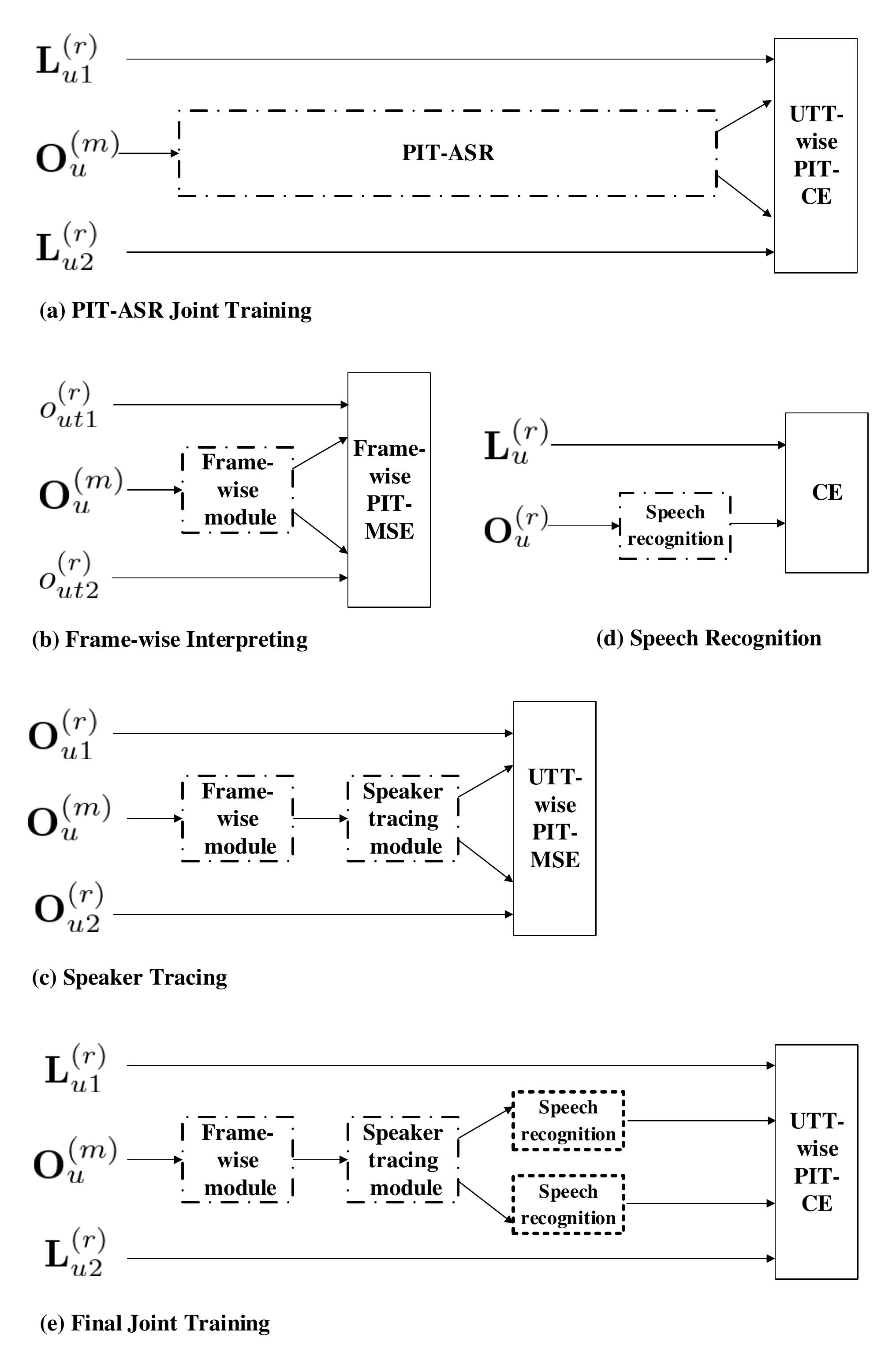}
    \caption{\it PIT-ASR Joint Training versus the  Modular Initialization and Progressive Joint Training. 
    The dash-dot blocks indicate the learnable model parameters. The dot-dot blocks indicate the learnable and shared model parameters.
    }
    \label{fig:modules}
\end{figure}

Although it is possible to train the modularized network of Figure~\ref{fig:modules}(e) from random initialization, it is better to use a progressive training strategy.
This strategy is motivated by the Curriculum learning theory in~\cite{bengio2009curriculum}, and integrates both modular initialization and joint training. We train a simple model first, and then use it as a pre-trained building block for a more complicated model and task. Figures~\ref{fig:modules}(b)-(e) illustrate how the model becomes progressively more complex while solving more difficult problems, from frame-wise mean squared error to whole utterance cross entropy.


Our simplest model, shown in Figure~\ref{fig:modules}(b), is trained to solve a frame-wise speech separation task.
For $N$ speakers, given the mixed data $\mathbf{O}_{u}^{(m)}$, the model infers an acoustic representation  ${o}_{utn}$ for each speaker $n$ at frame $t$ of utterance $u$. The objective function of  the frame-wise  training, is given as
\begin{equation}
\label{equ:fr-pit}
\begin{split}
\mathcal{J}_{\text{F-PIT}}=\sum_u\sum_t \frac{1}{N}\min_{s'\in \mathbf{S}} \sum_{n\in[1,N]} MSE({o}_{utn}^{(s')},{o}_{utn}^{(r)})
\end{split}
\end{equation}
where,  $\mathbf{S}$ is the permutation set of the reference representation and the inference representation. ${o}_{utn}^{(s')}$ and ${o}_{utn}^{(r)}$ is the frame level acoustic representation of permutation $s'$ and the reference clean speech, respectively. In each frame $t$ of the utterance $u$, the overall minimum square error,  $MSE$, is obtained by comparing all the reference and inference representations of each permutation $s'$.


The architecture for pre-training the speaker tracing module is explained in Figure~\ref{fig:modules}(c). The tracing module is combined with a pre-trained frame-wise module that has had its $N$ output layers removed.
As in \cite{yu2017permutation}, the PIT objective function is applied in utterance level.
\begin{equation}
\label{equ:utt-pit}
\begin{split}
\mathcal{J}_{\text{U-PIT}}=\sum_u \min_{s'\in \mathbf{S}} \sum_t \frac{1}{N} \sum_{n\in[1,N]} MSE({o}_{utn}^{(s')},{o}_{utn}^{(r)})
\end{split}
\end{equation}

The speech recognition module is separately pretrained in the same way as a conventional acoustic model, with clean speech and a cross-entropy objective function, maximizing $p(\mathbf{L}_u|\mathbf{O}_{u})$. This is illustrated in Figure~\ref{fig:modules}(d).

The final model, shown in Figure~\ref{fig:modules}(e), is created by stacking the speech recognition modules onto the outputs of the pre-trained speaker tracing and frame-wise modules.
It is jointly trained with an utterance level PIT-CE objective function given in Equation~(\ref{equ:ce-pit}).

Notably, even though the proposed structure has several copies of the speech recognition module, the numbers of parameters between Figure~\ref{fig:modules}(a) and Figure~\ref{fig:modules}(e) are similar. Because the speech recognition modules are solving similar problems, their parameters can be shared~\footnote{Namely the ``shared'' option in CNTK ``CloneFunction''.}. In preliminary experiments, the performance gap between sharing and independent parameters is less than 3\%. Thus to make the comparison fair, without specific explanation, the shared structure is used.



The advantage of the progressive joint training includes:
\begin{itemize}
\item Decreased model complexity leading to better system generalization and data efficiency. 
By separating system into proper modules, the model complexity is less than the all-in-one system in PIT-ASR. 
As unsupervised single-channel overlapped speech recognition   contains several of the hardest components in speech processing, the model complexity reduction is critical to the empirical training performance.
\item Faster convergence of the training process and better quality of the local minimum. Curriculum learning improves both the training speed and the performance of the model~\cite{bengio2009curriculum}. 
As shown in  Figure~\ref{fig:tr-curve}, the progressive joint training  needs  fewer epochs to 
converge, and it converges to a better local minimum.
An additional benefit is that the module initializations all   take much less time compared with the joint training~\footnote{The training curve of PIT speech separation can be referred to~\cite{yu2017permutation}. Besides, initializations also can be done in parallel.}. 
\item Potential to integrate with other technologies. 
State-of-the-art technologies in each field can be applied to the initialization of each module.
\end{itemize}


\subsection{Transfer Learning Based Joint Training}\label{Sec:self-transf}


Transfer learning, or teacher-student training, is a powerful technique to address domain adaptation problems in speech recognition. In this section, we show that multi-channel ASR is a type of domain adaptation, and that transfer learning can be used to improve model training.

\subsubsection{Conventional Transfer Learning in Domain Adaptation}\label{Sec:review-transf}

Transfer learning has been proposed to solve the distribution mismatch problem
in feature space~\cite{pan2010survey}.
To use this method in single-output ASR domain adaptation, parallel data must be available from a source domain and a target domain. A fully trained model in the source domain (the teacher) processes data and generates  posterior probabilities, which are sometimes referred to as ``soft labels.'' These soft labels then replace or augment the usual ``hard labels'' when training the student model with parallel data in the target domain~\cite{teastu2017li}.

To train the student, the Kullback-Leibler  divergence (KLD) between the output distributions of the teacher and student models is minimized as below. 
\begin{equation}
\label{equ:kld}
\begin{split}
KLD(y^{(T)},y^{(S)})=\sum_i y_i^{(T)} \log\frac{y_i^{(T)}}{y_i^{(S)}} \\
= \sum_i\ [\ y_i^{(T)} \log{y_i^{(T)}} - y_i^{(T)}\log{y_i^{(S)}}\ ]
\end{split}
\end{equation}
\begin{equation}
\label{equ:kld-f}
\begin{split}
= \sum_i\ - y_i^{(T)}\log{y_i^{(S)}}
\end{split}
\end{equation}
where $y_i^{(T)}$ and $y_i^{(S)}$ is the teacher  and student distributions respectively.
Because the first term is not related to the student model optimization, only the second term is used for optimization. Comparing Equation~(\ref{equ:kld-f}) to CE criterion in ASR, the hard labeling is replaced by the soft distribution inferred from the source data by the teacher model. 


\subsubsection{Self-transfer Learning}\label{Sec:review-transf}

In light of above discussion, self-transfer learning can be extended to the training of any multi-channel speech recognition system.
The student is, of course the multi-channel speech recognition system. It operates in the target domain of mixed speech acoustic data, and must produce separate outputs for each speaker in the mixture.
The teacher also must produce separate outputs for each speaker, but has access to the source domain: un-mixed clean speech. The teacher model is a set of clean speech acoustic models operating independently on the separate channels of clean speech.

\begin{figure}
  \centering
    \includegraphics[width=\linewidth]{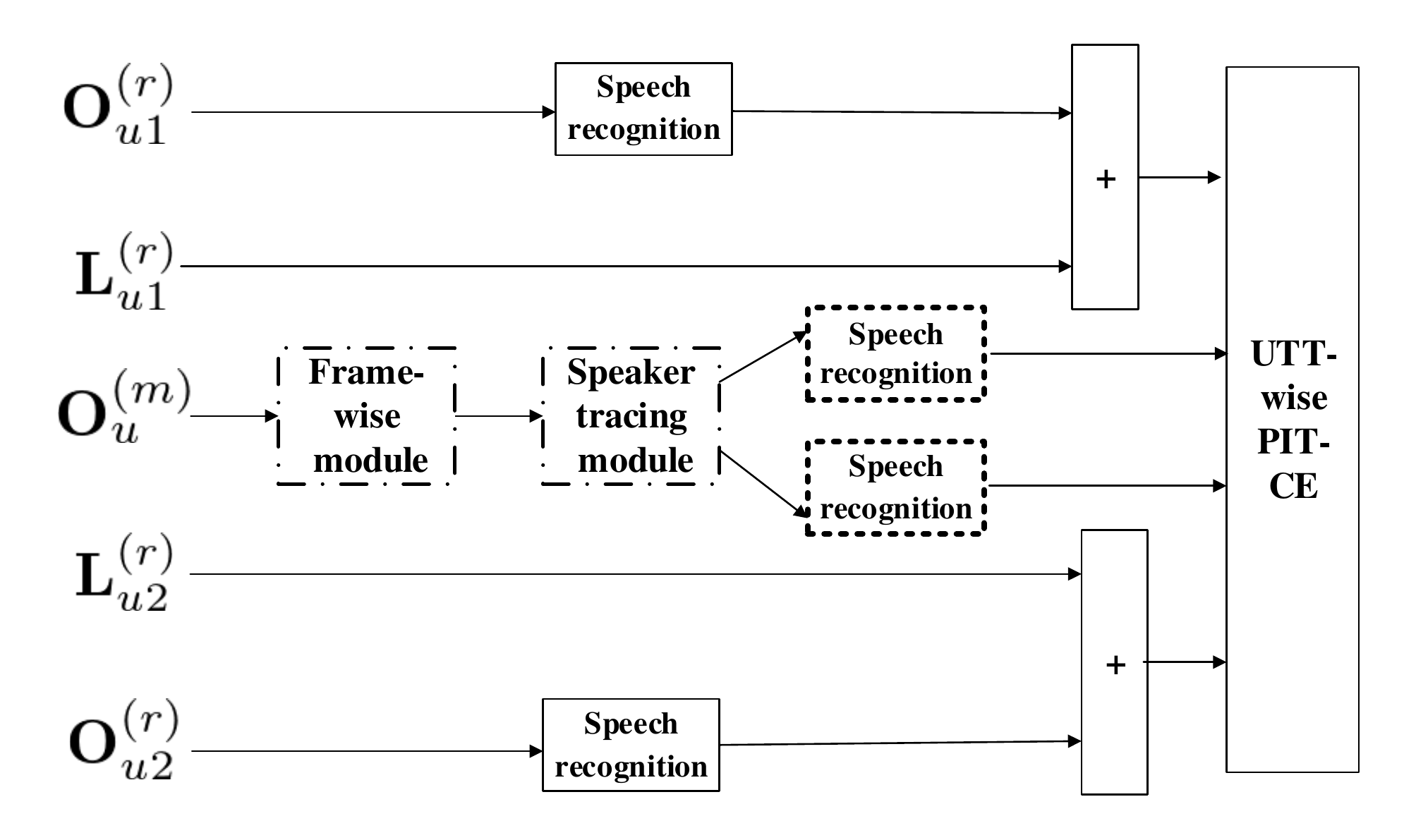}
    \caption{\it Transfer Learning Based Progressive Joint Training. The dash-dot blocks indicate the learnable model parameters. The dot-dot blocks indicate the learnable and shared model parameters.}
    \label{fig:joint-tr}
\end{figure}

The self-transfer learning method then minimizes the 
KLD between the output distribution of the mixed speech model and the set of clean speech models.
The KL divergence defined for utterance level PIT training between the clean speech model distribution and the joint model distribution  is as below,
\begin{equation}
\label{equ:kld-opt}
\begin{split}
\mathcal{J}_{\text{KLD-PIT}}=\sum_u \min_{s'\in \mathbf{S}} \sum_t \frac{1}{N} \sum_{n\in[1,N]} \\
KLD(P({l}_{utn}^{(c)}|\mathbf{O}_{un}^{(r)}),P({l}_{utn}^{(s')}|\mathbf{O}_{u}^{(m)}))
\end{split}
\end{equation}
where the calculation of  each $KLD(\cdot)$ pair is the same to the adaptation-purpose single-channel case in Equation~(\ref{equ:kld-f}).
Namely, the joint-trained model distribution, $y^{(S)}=P({l}_{utn}^{(s')}|\mathbf{O}_{u}^{(m)})$, is taken as the student model distribution,  and the clean speech model distribution, $y^{(T)}=P({l}_{utn}^{(c)}|\mathbf{O}_{un}^{(r)})$, is taken as the teacher model distribution.
It is notable that when this method is applied to the modular structure proposed in this work, as in Figure~\ref{fig:joint-tr}, the speech recognition modules can be initialized with an exact copy of the teacher model.

The training framework for  self-transfer learning is shown in Figure~\ref{fig:joint-tr}.
The soft targets generated by the teacher models are interpolated with the hard labeling as in~\cite{watanabe2017student}.
The training procedure is as below:
\begin{enumerate}
  \item Clone the speaker tracing layers in the bottom. Clone $2N$ copies of clean ASR model initialized in Section~\ref{Sec:modu-init}, half for stacking upon the speaker tracing layers, half for model inference given each clean speech stream.
  \item Use simultaneous clean speech streams $\mathbf{O}_{un}^{(r)}$ and the overlapped speech stream $\mathbf{O}_{u}^{(m)}$ to do joint training.
  \begin{enumerate}
  \item For each mini-batch, do forward propagation of the clean ASR model using each clean speech stream to calculate N streams of $P({l}_{utn}^{(c)}|\mathbf{O}_{un}^{(r)})$ respectively. Do forward propagation of the joint model using overlapped speech stream to calculate N streams of inference distributions, $P({l}_{utn}^{(s')}|\mathbf{O}_{u}^{(m)})$.
  \item For that mini-batch, calculate the error signal of Equation~(\ref{equ:kld-opt}) and then do back propagation for the joint model.
  \item Update parameters of the joint model and repeat until convergence.
  \end{enumerate}
\end{enumerate}

The proposed method elegantly solves the label mismatch problem and helps the model convergence.
Namely, 
using hard labeling obtained from forced-alignment in the clean speech is not proper, because the feature has been distorted in the mixed speech. The proposed method replaces it with the soft distribution.
In addition, 
the proposed method formulates the joint training of multi-channel ASR by domain adaptation between clean speech and overlapped speech.
Thus the soft distribution also helps model convergence, because it's easier to recreate its performance, compared with training a speech recognition model from scratch.

The evidence can be observed from the training curve in Figure~\ref{fig:tr-curve} that the initial CE of self-transfer learning based progressive joint training  is much better than that of both joint modeling and progressive joint modeling. 
Notably, the different starting points between the progressive joint modeling and self-transfer learning based progressive joint modeling is because the CE in the former system is calculated versus hard labeling, while for the latter system it is versus the soft distribution inferred from simultaneous clean speech~\footnote{We didn't use CE versus hard labeling for all systems, because CE between inference distribution and the hard labeling is not always a good indicator of the quality of the acoustic model~\cite{graves2013hybrid}, while the proposed curves better show the optimization process.}.
Thus with a better starting point and less parameter updating requirement, finally the model also comes into better minimum in the figure. 


The relationships of the proposed method and previous works are summarized as below.
\begin{itemize}[leftmargin=*] 
  \item Model space adaptation. The formulation of the proposed method is similar to KLD-based adaptation~\cite{yu2013kl} and teacher-student based domain adaptation~\cite{teastu2017li}. In \cite{yu2013kl}, to conservatively update model parameters using adaptation data,
  the target probability distribution is changed from the ground truth
  alignment to  a linear interpolation with the distribution
  estimated from the unadapted model. In \cite{teastu2017li},  
  the feature mismatch problem in the target domain is solved by minimizing the inference distribution divergence between the target and source domains using parallel-data.
  The reader is free to 
  consider  the proposed method as analogous to 
  optimizing student network in the target domain, i.e. overlapped speech, to behave similarly
  to the well-trained teacher network in the source domain, i.e. clean speech, while bearing in mind the proposed method requires module stacking because the motivation is to do joint training.
  
  \item Stereo piecewise linear compensation for environment (SPLICE)~\cite{deng2000large}.
  The SPLICE algorithm uses stereo data to do noise reduction and channel distortion compensation.
  In~\cite{markov2016robust}, the clean feature is used for the teacher model to provide supervision on the stereo noisy data trained student model. 
  In~\cite{watanabe2017student}, the multi-channel enhanced feature is used for the teacher model.
  In this work, the teacher-student framework is also based on stereo data. The student model is initialized better to cope with the more difficult modeling problem, and the entire framework is expanded to handle multiple output streams. 
  
  \item Progressive stacking transfer learning.
  \cite{wang2017transfer} proposes to progressively conduct transfer learning to train speech enhancement layers. The motivation of the progressive stacking is only to gradually model a hard task by dividing into several same but smaller tasks.
  Thus the criteria of all   tasks are  the same.
  However, the proposed method is to do joint training of distinct tasks. And  each task is fully trained with specific data and criterion.

  \item Self-supervised training.
  \cite{hadsell2009learning} proposes to use  a teacher model based on more accurate sensor information as the supervision of the student model.
  The motivation of the proposed method is different, which is to transfer distribution between two models with feature mismatch. Besides, the empirical procedure of the proposed method is to fine-tune the original model in the target feature distribution from supervision of itself in parallel source distribution.

\end{itemize}

\subsubsection{Learning from Ensemble}\label{Sec:learn-ensem}

Following the transfer learning diagram, the joint model can also benefit from an ensemble of teachers~\cite{ba2014deep}. Specifically, because the problem includes several sub-problems discussed in Section~\ref{Sec:modu-init}, different neural network structures can show different superiorities, e.g., with different numbers of stream-dependent layers and stream-independent layers. Learning from an ensemble of joint models with different structures is promising in both performance improvement and model compression.

\subsection{Multi-output Sequence Discriminative Training}\label{sec:seq-disc-tr}


\subsubsection{Motivation}
Speech recognition is inherently a sequence prediction problem. In single-output ASR, sequence level criteria such as sequence discriminative training tend to improve performance.
The unsupervised single-channel overlapped speech recognition problem further includes the speaker tracing problem, which is also a sequence level problem. Previous works concatenate frame level
CE as the criteria of the sequence level problem, which
limits the modeling effect of the neural network.
In this paper, sequence discriminative  training of multiple output streams is  proposed for the first time.

\subsubsection{Sequence Discriminative Criterion of Multiple outputs}
In single-output ASR, to form a sequence discriminative training criterion, it is necessary to calculate the sequence posterior probability using Bayes' theorem as below,

  
\begin{equation}
\label{equ:single-mbr}
\begin{split}
P(\mathbf{L}_u|\mathbf{O}_u)=\frac {p(\mathbf{O}_u|\mathbf{L}_u)P(\mathbf{L}_u)}{p(\mathbf{O}_u)}  
\end{split}
\end{equation}
Here, $\mathbf{L}_u$ is the word sequence of utterance $u$. $P(\mathbf{L}_u)$ is the language model probability.
$p(\mathbf{O}_u|\mathbf{L}_u)$ is the corresponding acoustic part.
The marginal probability $p(\mathbf{O}_u)$ of the feature sequence $\mathbf{O}_u$, is modeled by summation of the probability over all possible hypothesis sequences. 
\begin{equation}
\label{equ:po-prob}
\begin{split}
p(\mathbf{O}_u)=\sum_\mathbf{L} p(\mathbf{O}_u,\mathbf{L})= \sum_\mathbf{L} P(\mathbf{L}) p(\mathbf{O}_u|\mathbf{L})
\end{split}
\end{equation}
Here, $\mathbf{L}$ denotes all competing hypotheses. 
As an example of the sequence discriminative training criteria, the maximum mutual information (MMI)~\cite{vesely2013sequence} of inference distribution stream $\mathbf{L}_u$ in utterance $u$ is defined as below,

\begin{equation}
\label{equ:single-mmi}
\begin{split}
\mathcal{J}_{\text{SEQ}}(\mathbf{L}_u,\mathbf{L}_u^{(r)})= \log P(\mathbf{L}_u^{(r)}|\mathbf{O}_u)
\end{split}
\end{equation}
where $\mathbf{L}_u^{(r)}$ is the corresponding reference.

For the overlapped speech recognition problem, the conditional independence assumption in the output label streams is still made as in Equation~(\ref{equ:obj-joint}). Then the cross-entropy based PIT can be transformed to sequence discriminative criterion based PIT as below,

\begin{equation}
\label{equ:bayes-pit}
\begin{split}
\mathcal{J}_{\text{SEQ-PIT}}=\sum_u \min_{s'\in \mathbf{S}} \frac{1}{N} \sum_{n\in[1,N]}-\mathcal{J}_{\text{SEQ}}(\mathbf{L}_{un}^{(s')},\mathbf{L}_{un}^{(r)})
\end{split}
\end{equation}
Different from Equation~(\ref{equ:ce-pit}), the best permutation is decided by $\mathcal{J}_{\text{SEQ}}(\mathbf{L}_{un}^{(s')},\mathbf{L}_{un}^{(r)})$, which is the sequence discriminative criterion of taking the $s'$-th permutation in $n$-th output inference stream  at utterance $u$. Similar to CE-PIT, $\mathcal{J}_{\text{SEQ}}$ of all the permutations are calculated and the minimum permutation is taken to do the optimization.

The definition of  $\mathcal{J}_{\text{SEQ}}(\mathbf{L}_{un}^{(s')},\mathbf{L}_{un}^{(r)})$ is similar to Equation~(\ref{equ:single-mmi}) in single output ASR. 
\begin{equation}
\label{equ:lf-mmi}
\begin{split}
\mathcal{J}_{\tt{MMI}}
=\sum_u \mathcal{J}_{\text{SEQ}}(\mathbf{L}_{un}^{(s')},\mathbf{L}_{un}^{(r)}) \\
=\sum_{u} \log \frac {\sum_{\mathbf{L}_u} p(\mathbf{O}_u|\mathbf{L}_u)^{\kappa}P(\mathbf{L}_u)}{\sum_{\mathbf{L}} p(\mathbf{O}_u|\mathbf{L})^{\kappa}P(\mathbf{L})}  
\end{split}
\end{equation}
For simplicity, $\mathbf{L}_u=\mathbf{L}_{un}^{(r)}$ and $\mathbf{L}$ is  all the hypothesis sequences generated by the output stream $\mathbf{L}_{un}^{(s')}$.
$p(\mathbf{O}_u|\mathbf{L}_u)$ and $p(\mathbf{O}_u|\mathbf{L})$ is the conditional
likelihood obtained from forward propagation of the joint model, $P(\mathbf{L}_u)$ and $P(\mathbf{L})$ is the prior probability obtained from language model.

\subsubsection{Competing Hypothesis Modeling}
\label{sec:disc-tr-emp}

The hypothesis sequence $\mathbf{L}$ refers to all the competing hypotheses in the speech recognition. Bad modeling  of $\mathbf{L}$, namely ignoring some modeling errors, results in  imprecise estimation of Equation~(\ref{equ:po-prob}), which hurts ASR performance.
Thus competing hypotheses modeling is key to the discriminative training. Empirically, 
it is constrained by linguistic search space. In the single output ASR, the linguistic search space is further pruned by the online method, lattice-based discriminative training~\cite{povey2005discriminative}, or by the offline method, lattice-free (LF) discriminative training~\cite{chen2006advances,povey2016purely,xiong2016achieving}.

Compared with single output  ASR, the additional  error hypothesis types
include: i) Assignment errors: someone said a word, but it is assigned to the wrong channel. ii) cross talk errors: one person said a word, but it appears in multiple channels. 
They both come from imperfect acoustic modeling and result in several challenges in discriminative training. 

The first problem is linguistic search space modeling. As discussed in the first paragraph, there are mainly two branches of methods. 
When training the joint model, speaker tracing results can always change, which results in different permutations  of the same utterance between different epochs.
Thus if using lattice-based method, lattices should be updated after each epoch in case of bias in the search space modeling. Another choice is to use a pre-pruned senone level language model as the common search space for all utterances~\cite{xiong2016achieving}. With this method, the lattice generation problem in multiple outputs can be solved and the discriminative training can be conducted efficiently in the shared search space.

The second problem is the swapped word modeling in multiple outputs.
Swapped word results in both cross talk errors and assignment errors. Thus bad swapped word modeling hurts the ASR performance.
Generally, the linguistic search space is estimated from the transcription of the training dataset. And then sequence criterion is calculated in this search space. 
Because there's no swapped word phenomenon in the transcription, thus the search space doesn't contain swapped word cases, which results in overestimating the sequence criterion. Especially for the multiple output streams, the swapped word errors are critical to the ASR performance. Three methods are proposed to cope with the problem.
\begin{itemize}[leftmargin=*]
  \item Artificial swapped words.
A very simple method is to generate several copies of the transcription with artificially swapped words in each copy. And then the language model is estimated on the re-generated transcription. With this method, some of the swapped word cases can still exist in the search space. Thus the problem is alleviated. Empirically, the senone level language model is obtained from the senone level transcription, namely clustered tri-phone state alignment.
In case of significant increasing in the search space because of the swapped word, we set a rule that in each frame, the  probability of senone swapping is $\alpha$. But if the senone is swapped, the senone sequence of the  following $\beta$ frames won't be swapped. And $\gamma$ copies of the transcriptions are generated.
  \item De-correlated lattice free MMI (LF-DC-MMI).
The motivation is that swapped words come from the other output streams. Thus adding these output streams into the search space and minimizing them in the denominator of discriminative training can alleviate the problem.
\begin{equation}
\label{equ:lf-dc-mmi}
\begin{split}
\mathcal{J}_{\tt{LF\text{-}DC\text{-}MMI}}
=\sum_{u} \log  [ \frac {\sum_{\mathbf{L}_u} p(\mathbf{O}_u|\mathbf{L}_u)^{\kappa}P(\mathbf{L}_u)}{(\ \sum_{\mathbf{L}} p(\mathbf{O}_u|\mathbf{L})^{\kappa}P(\mathbf{L})\ )^{1-\lambda} } 
\cdot \\ 
\frac{1} {(\ {\sum_{\mathbf{L}_{\hat{u}}}} p(\mathbf{O}_u|{\mathbf{L}_{\hat{u}}})^{\kappa}P({\mathbf{L}_{\hat{u}}})\ )^\lambda}
 ]
\end{split}
\end{equation}
In Equation~(\ref{equ:lf-dc-mmi}), the other output streams are denoted as $\mathbf{L}_{\hat{u}}$. An interpolation weight $\lambda$ is added with the augmented term in the denominator.

  \item De-correlated lattice free boosted MMI (LF-DC-bMMI).
Analogous to boosted MMI~\cite{povey2008boosted} as Equation~(\ref{equ:lf-bmmi}), 
\begin{equation}
\label{equ:lf-bmmi}
\begin{split}
\mathcal{J}_{\tt{LF\text{-}bMMI}}
=\sum_{u} \log \frac {\sum_{\mathbf{L}_u} p(\mathbf{O}_u|\mathbf{L}_u)^{\kappa}P(\mathbf{L}_u)}{\sum_{\mathbf{L}} p(\mathbf{O}_u|\mathbf{L})^{\kappa}P(\mathbf{L})e^{-b\ \mathop{\max}_{\mathbf{L}_u} A(\mathbf{L},\mathbf{L}_u)}}  
\end{split}
\end{equation}
we propose de-correlated lattice free boosted MMI (LF-DC-bMMI) as Equation~(\ref{equ:lf-dc-bmmi}).
Here, $b$ is the boosting factor. $A(\mathbf{L},\mathbf{L}_u)$ is the state level accuracy between sequence $\mathbf{L}$ and $\mathbf{L}_u$. By this method, the ASR error hypotheses can be further minimized in the denominator. 
In the proposed method, both the ASR errors between the target inference sequence and the target reference, and the falsely-recognition of the interfere streams, are boosted. 
\begin{equation}
\label{equ:lf-dc-bmmi}
\begin{split}
\mathcal{J}_{\tt{LF\text{-}DC\text{-}bMMI}}
=\sum_{u} \log\ [\ \sum_{\mathbf{L}_u} p(\mathbf{O}_u|\mathbf{L}_u)^{\kappa}P(\mathbf{L}_u)\cdot \\
\frac {1}{\sum_{\mathbf{L}} p(\mathbf{O}_u|\mathbf{L})^{\kappa}P(\mathbf{L})e^{-b\ \mathop{\max}_{\mathbf{L}_u} A(\mathbf{L},\mathbf{L}_u)
-\hat{b}\ {\mathop{\max}_{\mathbf{L}_{\hat{u}}}} 
(1-A(\mathbf{L},\mathbf{L}_{\hat{u}}) )
}}\ ]
\end{split}
\end{equation}
where $\hat{b}$ is the de-correlated boosting factor and $A(\mathbf{L},\mathbf{L}_{\hat{u}})$ measures how many falsely recognitions of  the interfere streams.
\end{itemize}

Experiments are conducted on all three methods in Section~\ref{Sec:exp-disc-tr} and the first method can be further combined with the other two.
%
%


\section{Experiment}\label{Sec:exp}
The experimental results
are reported in artificial overlapped Switchboard corpus and
Eval2000 hub5e-swb test set. 
Although the methods presented here are valid for any number of overlapped speakers, we focus on the two-talker scenario.

\subsection{Experimental Setup}\label{Sec:}

For training, the Switchboard corpus~\cite{godfrey1992switchboard} is used, which contains about
300 hours of speech.
Evaluation is carried out on the Switchboard (SWB) subset of the NIST
2000 CTS (hub5e-swb) test set. The waveforms were segmented
according to the NIST partitioned evaluation map (PEM) file. 

Two-talker overlapped speech is artificially generated by mixing these waveform segments.
To maximize the speech overlap, we developed a procedure to mix similarly sized segments at around 0dB.
First, we sort the speech segments by length.
Then, we take segments in pairs, zero-padding the shorter segment so both have the same length.
These pairs are then mixed together to create the overlapped speech data.
The overlapping procedure is similar to~\cite{yu2017recognizing} except that we make no modification to the signal levels before mixing~\footnote{We will try to release  scripts that create the training data  in the future.}.
%
%
After overlapping, there's 150 hours data in the training, called {\em{150 hours dataset}}, and 915 utterances in the test set. After decoding, there are 1830 utterances for evaluation, and the shortest utterance in the hub5e-swb dataset is discarded.
Additionally, we define a small training set, the {\em 50 hours dataset}, as a random 50 hour subset of the {\em 150 hours dataset}.
 Results are reported using both datasets.  

In the training stage, 
80-dimensional log-filterbank features were extracted every 10 milliseconds, using a 25-millisecond analysis window.
The convolution neural network (CNN) models use 41 context frames (20 in both left and right)
and the long short term memory networks (LSTM) processed one frame of input at a time. All neural networks were trained with the Microsoft Cognitive Toolkit (CNTK)~\cite{seide2016cntk}.
The detailed setup of CNN is listed in Section~\ref{Sec:exp-modular}.
The acoustic model is based on three state left-to-right triphone models with 9000 tied states (senones). The individual senone alignments for the two-talkers in each mixed speech utterance
are from the single-speaker ASR alignment~\cite{xiong2016achieving}. For compatibility, the alignment of the shorter utterance within the mixed speech is padded with the silence state at the front and the end.
The clean speech recognition performance in the corpus can be referred to \cite{xiong2016achieving,povey2016purely}. Using clean speech model to do decoding in the overlapped speech isn't reported as it's  as bad as in~\cite{yu2017recognizing}.
The baseline model of joint training is a PIT-ASR model with a setup similar to~\cite{yu2017recognizing}. The PIT-ASR model is composed of 10 bidirectional LSTM layers with 768 memory cells in each layer~\footnote{In the updated arxiv version of~\cite{yu2017recognizing}, the number of layers changed from 10 to 4 and  a moderate WER improvement is obtained. Since the authors didn't provide further information, this work mainly compares with the result of 10 layers. In this work, more layers always bring about better performance. And we believe that improved performance with fewer layers in~\cite{yu2017recognizing} is a result of bad model generalization in the PIT-ASR model framework.}, and 80-dimensional feature. 
The baseline model of separately optimized system is a PIT for speech separation (PIT-SS) model combined with a  clean speech  ASR model. As PIT-SS model has shown competitive performance compared with other speech separation systems, only PIT-SS model is taken as the baseline. The PIT-SS model has a setup similar to~\cite{yu2017permutation} but  with 6  bidirectional LSTM layers with 768 memory cells in each layer, it directly 
outputs multiple channels of the 80 dimensional log-filterbank features the speech recognition module expects.
The speech recognition module, pretrained as a clean speech model,  is  composed of 4 bidirectional LSTM layers with 768 memory cells in each layer.
It is trained from the corresponding source speech 
segments used to create the overlapped corpus.
After initialization, the WER performance of the ASR model in the clean speech test set is 17.0\%. Although it would be easy to incorporate a stronger acoustic model~\cite{xiong2016achieving} in conjunction with the proposed method, we chose a structure that allows for a fair comparison, in terms of the number of model parameters, among the baselines and proposed methods. However, as discussed in Section~\ref{Sec:modu-init}, the modular system needs fewer parameters and training iterations to achieve good performance. 
Notably, the use of enhanced signals after speech separation as training data of the speech recognition module tends to degrade the ASR performance and isn't included. 
The reason can be from the sporadic distortions that signal processing inevitably adds,  similar to what has been observed in~\cite{watanabe2017student}. 

In the evaluation stage, a 30k-vocabulary language model derived from the most common words in the Switchboard and Fisher corpora is used. 
The decoder uses a statically compiled unigram graph, and dynamically applies the language model score. The unigram graph
has about 300k states and 500k arcs~\cite{xiong2016achieving}. 
Two outputs of the PIT-ASR model are both used in decoding to obtain the hypotheses for two talkers. For scoring, we
evaluated the hypotheses on the pairwise score mode against
the two references, and used the assignment with better word error rate (WER)  for each utterance~\cite{yu2017recognizing}.
Only the average WER of two output streams is reported, 
as the task is to correctly recognize all words from both speakers.

\subsection{Separate Optimization v.s. Joint Modeling}\label{Sec:exp-modular}

Table~\ref{tab:sep-joint} shows the performance of the naive joint modeling, PIT-ASR, compared with that of the separately optimized system.

\begin{table}[thbp!]
  \caption{\label{tab:sep-joint} {\it  Separate Optimization v.s. Joint Modeling. Fine-tune ST denotes to fine-tune the speaker tracing module and Fine-tune ASR denotes to fine-tune the speech recognition module defined in Figure~\ref{fig:modules}.}}
  \centerline{
    \begin{tabular}{c c c c||cc}
      \hline
      Layers  &Modular &Fine-tune ST&Fine-tune ASR&WER&Rel. (\%)\\
      \hline \hline
      10 $\cdot$ 0 & $\times$ & n/a & n/a & 57.5 & 0\\
      \hline\hline
      \multirow{4}*{ 6 $\cdot$ 4} & $\times$& n/a & n/a  & 52.8 & -8.2\\
      \cline{2-6}
        & $\surd$ & $\times$ & $\times$ & 93.4&+62.4 \\
      & $\surd$&  $\surd$ &$\times$  & 51.3 & -10.7 \\
      & $\surd$ & $\surd$ & $\surd$ & 50.2 &-12.7\\
      \hline
    \end{tabular}
  }
\end{table}

The first row shows the performance of the joint training baseline model in this corpus, PIT-ASR~\cite{yu2017recognizing}. Compared with the 0dB WER result listed in~\cite{yu2017recognizing}, 55.80\%, the performance is reasonable~\footnote{The differences include: the dataset is 50 hours versus 400 hours, while the clean speech WER is 17.0\% versus 26.6\% . In Section~\ref{Sec:exp-150hrs}, the experiments are extended to 150 hours corpus.}.
As discussed in Section~\ref{Sec:modu-init}, the separately optimized system has a similar number of parameters but different model architecture. Thus to make a fair comparison, the model with 6 stream-independent layers in the bottom and 4  parameter-shared stream-dependent layers in the top, denoted as 6$\cdot$4, is listed in the second row~\footnote{As a comparison, the baseline 10$\cdot$0 has 10 stream-independent layers in the bottom and no stream-dependent layers except the output layers.}. The learnable structure is the same to the dash-dot  blocks shown in Figure~\ref{fig:modules}(e), but trained from scratch as 10$\cdot$0. 
The performance of the 6$\cdot$4 structure is significantly better than that of the 10$\cdot$0 structure. The reason is that unlike in the pure speech separation task, the speech recognition stage in this task  is also very hard and needs more  nonlinear layers. It also shows that this task is much harder than the speech separation, so better joint training method to fulfill the performance of each module is critical to the success.

Without fine-tuning  parameters, the performance of the separately optimized system is shown in the third row. The significantly worse performance comes from the feature mismatch in Equation~(\ref{equ:obj-sep}).
With fine-tuning parameters, the performance is restored in the fourth and fifth rows. The system in the fifth row can be viewed as a strong baseline with separate optimization and fine-tuning.

\begin{figure}
{
  \centering
    \includegraphics[width=\linewidth]{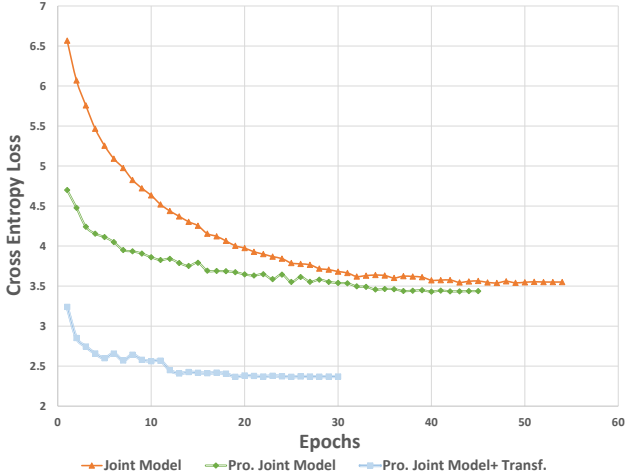}
    }
    \caption{\it Validation Curves of Naive Joint Modeling and the Proposed Methods. Joint modeling, progressive joint modeling and transfer learning based progressive joint modeling are denoted as Joint Model,  Pro. Joint Model and Pro. Joint Model + Transf. in the figure. Each epoch contains 24 hours of data. } 
    \label{fig:tr-curve}
\end{figure}

The better performance of the progressive joint training is from better model generalization and training efficiency~\footnote{After submitting the paper, we notice that  similar progressive joint training idea is proposed in \cite{qian2017single}. Experiments using the AMI corpus~\cite{hain2012transcribing} shows similar improvement.}. 
Figure~\ref{fig:tr-curve} show the effect.
Training curves of both joint modeling, i.e. the second row in the table, and progressive joint modeling, i.e. the fifth row in the table, are plotted.
From the figure, both better starting point and better converged minimum can be observed in the joint progressive training. 
With better joint training strategy shown in Section~\ref{Sec:exp-self-joint}, such modeling effect can be further fulfilled.

Table~\ref{tab:modular} shows the performance of the joint modeling from different modular initialization setups. All modular systems are fine-tuned after joint training.
The first and second rows show the naive joint training models with BLSTM and BLSTM combined with CNN, respectively.
{6$\cdot$4 BLSTM} refers to 6 layers BLSTM in the bottom and 4  parameter-shared layers for each output in the top, i.e.  6 $\cdot$ 4 in Table~\ref{tab:sep-joint}.
Layer-wise context expansion with attention (LACE) model is used for CNN~\cite{yu2016deep}, which is a TDNN~\cite{waibel1989phoneme} variant where each higher layer is a weighted sum of nonlinear transformations of a window of lower layer frame. Each LACE block starts with a convolution layer with stride 2 which sub-samples the input and increases the number of channels. This layer is followed by three  RELU-convolution layers with jump links.  The channel size is 48. 
The slightly different setup compared with~\cite{xiong2016achieving} is to make  parameter number of one LACE block comparable with one layer of bidirectional LSTM with 768 memory cells, i.e. 5M parameters.
Only one block of LACE is used to replace one layer of BLSTM as the frame-wise interpreting module, because the speaker tracing and speech recognition modules are the hardest parts in the problem. The other parts of the structure are the same, and the proposed structure is denoted as {1 LACE + 5$\cdot$4 BLSTM}.
From the table, it can be observed that there's no improvement by  merely stacking these kinds of neural networks together and jointly training them.

In the third and fourth rows, the model is firstly initialized with frame-wise interpreting, Figure~\ref{fig:modules}(b), speaker tracing,  Figure~\ref{fig:modules}(c), and speech recognition,  Figure~\ref{fig:modules}(d), tasks respectively and then jointly trained. Comparing the fourth row to the third row, {1 LACE + 5$\cdot$4 BLSTM} shows larger improvement than {6$\cdot$4 BLSTM}. Two conclusions can be derived from the results: i) CNN structure is more suitable for the frame-wise interpreting module because it focuses on the local context and has better modeling power of frequency variations~\cite{sainath2015convolutional}. 
Meanwhile, LSTM is good at temporal modeling, which is more suitable for the speaker tracing and speech recognition module. The architecture with  {1 LACE + 5$\cdot$4 BLSTM} layers combines their strength. Notably, \cite{sainath2015convolutional} proposes a similar structure, called CLDNN, to form the acoustic model with modules focusing on different scales and take advantage of the complementarity of CNN, LSTM and DNN. The difference is that, to fulfill the respective advantages in modeling, the proposed method further pretrains each module with different criteria. ii) As the performance improvement from modularization and initialization is much larger in {1 LACE + 5$\cdot$4 BLSTM}, it shows that 
module-wise initialization is important to fulfill the modeling power of neural networks especially with different  structures and scales.  

To further analyze the frame-wise interpreting ability of CNN and BLSTM, experiments without frame-wise interpreting initialization are conducted in the fifth and sixth rows. This time, the performances are similar both in {6$\cdot$4 BLSTM} and {1 LACE + 5$\cdot$4 BLSTM}. It shows that in initialization of the speaker tracing module, both BLSTM and CNN can spontaneously learn the frame-wise interpreting ability. 
We  notice that in~\cite{yu2017permutation}, frame-wise PIT training doesn't show good performance, which is similar to our observation. 


\begin{table}[thbp!]
  \caption{\label{tab:modular} {\it  Progressive Joint Modeling Based on Modular Initialization. Initializations of frame-wise interpreting, speaker tracing and speech recognition are denoted as FI, ST and ASR respectively.}}
  \centerline{
    \begin{tabular}{c c ||c c}
      \hline
      Modular Init. &
      Neural network  &
      WER&
      Rel. (\%)\\
      \hline \hline
      \multirow{2}*{$\times$} & {6$\cdot$4 BLSTM}  & 52.8 & 0 \\
       & {1 LACE + 5$\cdot$4 BLSTM}  &  52.9& +0.2 \\
       \hline
      \multirow{2}*{FI+ST+ASR} & {6$\cdot$4 BLSTM}  & 50.3 & -4.9 \\
      & {1 LACE + 5$\cdot$4 BLSTM}  & 47.4 & -10.2 \\
      \hline
      \multirow{2}*{ST+ASR}  & {6$\cdot$4 BLSTM}  & 50.2 & -5.0 \\
      & {1 LACE + 5$\cdot$4 BLSTM}  & 47.4 & -10.2 \\
      \hline
    \end{tabular}
  }
\end{table}


\subsection{Self-transfer Learning Based Joint Modeling}\label{Sec:exp-self-joint}

Table~\ref{tab:trans-joint} shows the performance improvement of the transfer learning applied to joint modeling. For transfer learning, the interpolation weight between hard and soft labels is 0.5 . 

\begin{table}[thbp!]
  \caption{\label{tab:trans-joint} {\it  Transfer Learning Based Joint Modeling}}
  \centerline{
    \begin{tabular}{c c c||c c}
      \hline
      Layers &
      Modular &
      teacher &
      WER&
      Rel. (\%)\\
      \hline \hline
      \multirow{3}*{10$\cdot$0} & $\times$ &  $\times$ &  57.5 & 0 \\
       & $\times$ & 9$\cdot$1 $\oplus$ 6$\cdot$4 $\oplus$ 3$\cdot$7 & 55.0 & -4.4 \\
       & $\times$ & \multirow{1}*{ clean } & 52.5 &-8.7 \\
    \hline  \hline
      \multirow{4}*{6$\cdot$4} & $\times$ &  $\times$ & 52.8 & -8.2\\
      & $\times$ &   clean  & 47.1& -18.0 \\
       & $\surd$ &  clean & 38.9 & -32.4\\
       & $\surd$ & MMI clean  & 35.8 &-37.7 \\
      \hline
    \end{tabular}
  }
\end{table}

The original PIT-ASR system is in the first row and a better PIT-ASR baseline with 6$\cdot$4 structure in Table~\ref{tab:sep-joint} is also included in the fourth row. 

The ensemble-based transfer learning proposed in Section~\ref{Sec:learn-ensem} is  tested in the second row. The ensemble contains 3 types of  structure,  9$\cdot$1, 6$\cdot$4 and 3$\cdot$7, where the left number denotes the bottom stream-independent layers and the right number denotes the top stream-dependent layers. The student network learns from each teacher one-by-one. Although it's not a large gain, it shows improvement after learning from each teacher.

The result of replacing hard labeling with  simultaneous clean speech based transfer learning is listed in the third and the fifth rows. In both model architectures, transfer learning brings about a relative 10\% improvement over the respective baseline, which is comparable with the result in~\cite{teastu2017li}. It shows that soft distribution inferred by the model with similar architecture is 
superior to the 
hard labeling.

The self-transfer learning based progressive joint modeling is finally listed in the sixth and the seventh row by using CE-trained and MMI-trained clean speech teacher respectively. Notably,  as the model framework discussed in~\ref{Sec:self-transf}, the initializations of speech recognition modules are the respective clean speech teachers. The result shows over 30\% relative improvement. Comparing the third and the fifth rows with Table~\ref{tab:modular}, it can be further observed that, combining progressive joint training  and self-transfer learning brings about even larger improvement compared with the summation of the relative improvement from each of the two technologies. The learning curve of the proposed method is also shown in Figure~\ref{fig:tr-curve}.

From these results, we conclude: 
i)  The proposed method brings about faster convergence and  better converged minimum. The reason is discussed in Section~\ref{Sec:self-transf}. 
The better convergence result also comes from  the removal of the inappropriate hard alignment in the joint training.
ii) Easier convergence helps the model fulfill the best performance in each module. That's the explanation of  the even better synergy result compared with the summation of the relative improvements from transfer learning and progressive joint training.
iii) Better teacher generates better student. And  the MMI-trained distribution can also be transferred to the student model, similar to what has been observed in~\cite{li2014learning}.

\begin{figure}
  \centering
    \includegraphics[width=\linewidth]{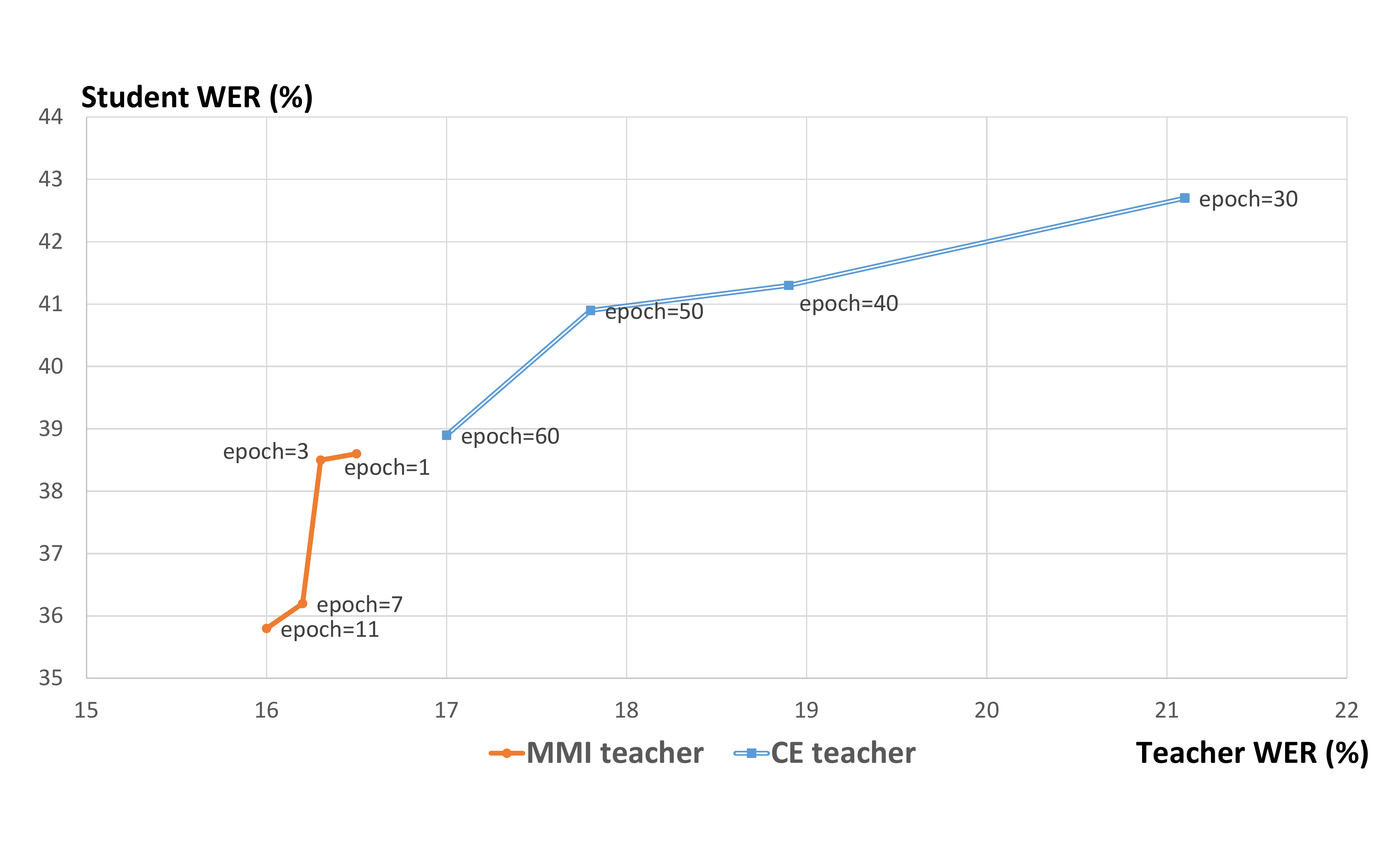}
    \caption{\it Student Performance versus Teacher Quality in Transfer Learning Based Progressive Joint Modeling. All the student models are fully converged based on the respective teacher model. The epoch number denotes how many epochs the teacher model is trained in the clean speech. Each epoch contains 24 hours of data. The MMI teachers are initialized from the best CE teacher, i.e. epoch=60 .}
    \label{fig:tea-wer-stu-wer}
\end{figure}

Figure~\ref{fig:tea-wer-stu-wer} further shows the student performance versus quality of the teacher in transfer learning  based joint modeling.
It can be observed that better student can usually be obtained  with better teacher. An important reason is that self-transfer learning is conducted by minimizing the divergence of its own distributions in mixed speech and clean speech. Thus better original distribution, including MMI-trained distribution, can intrinsically be part of the joint model and brings about better results. The only inflection point is in epoch=3 of the MMI teacher, where the student performance is similar to epoch=1 although the teacher model has better WER performance. We believe the reason is that the distribution of the teacher model of epoch=3 is hard to transfer to the student model because of the transition process from CE-trained distribution to  MMI-trained distribution.

\subsection{Sequence Discriminative Training}\label{Sec:exp-disc-tr}

Table~\ref{tab:disc-tr} shows the performance improvement of sequence discriminative training based joint modeling. In this experiment, the baseline is PIT-ASR model without progressive joint training. All the structures are kept the same as 10$\cdot$0 and only criteria are changed. $\kappa=0.1$, which is in accordance with the decoding setup.
The senone level language model for competing hypothesis modeling is obtained from the clustered tri-phone state alignment. Tri-gram is used, similar to~\cite{xiong2016achieving}.
For the artificial swapped word method, the hyper-parameters in Section~\ref{sec:disc-tr-emp} is decided by the preliminary experiment. $\alpha=0.4$, $\beta=10$ and $\gamma=2$. The swapped word senone level search graph is 4 times larger than the original tri-gram senone level language model graph.  $\lambda=0.1$, $b=0.1$ and $\hat b=0.2$. 

\begin{table}[thbp!]
  \caption{\label{tab:disc-tr} {\it  Sequence Discriminative Training Based Joint Modeling}}
  \centerline{
    \begin{tabular}{c c c||c  c}
      \hline
      corpus &
      criterion &
      Den. graph&
      WER &
      Rel. (\%)\\
      \hline \hline
      \multirow{7}*{50 hours} & PIT-CE  & n/a &57.5& 0 \\
       \cline{2-5}
      &LF-MMI &\multirow{4}*{senone transcription} & 54.6 &  -4.9\\
      & LF-DC-MMI & &53.2 &-7.3\\
       &LF-bMMI & & 53.5 & -6.9\\
       &LF-DC-bMMI && 53.1&-7.4\\
       \cline{2-5}
      &LF-MMI & \multirow{2}*{+ art. swapped word}   & 53.5 &-6.9 \\
       & LF-DC-bMMI&  &52.7 & -8.2\\
      \hline\hline
       \multirow{3}*{150 hours} & PIT-CE  & n/a &42.2& 0 \\
       \cline{2-5}
       &LF-MMI &\multirow{2}*{senone transcription} & 40.1 & -4.9 \\
       &LF-DC-bMMI  & & 39.1 & -7.3 \\
       \hline
    \end{tabular}
  }
\end{table}

The baseline PIT-ASR system is shown in the first row, denoted as PIT-CE to show the criterion of the system. Applying the naive sequence discriminative training method for multiple outputs in the second row, only brings about 4.9\% relative improvement. For the recognition result, word precision is improved but insertion error increases. The reason is from imperfect search space modeling of swapped words discussed in~\ref{sec:seq-disc-tr}. 

By applying the proposed LF-DC-MMI method in the third row, the recognition result is significantly improved compared with both baseline and the naive LF-MMI.
The proposed method minimizes the swapped words from parallel output streams in the denominator modeling. Thus the problem can be alleviated.

The fourth and fifth rows show the effect of using bMMI instead of MMI in the formulations.
The bMMI criterion boosts the ASR errors in the denominator modeling, implicitly including possibly swapped words. 
Although significant improvement can be observed between LF-MMI and LF-bMMI, the LF-DC-bMMI explicitly includes the swapped words and achieves an even better result.

The proposed artificial swapped word method is shown in the sixth and seventh rows. 
By comparing the sixth row with the second row, and comparing the seventh row with the fifth row, it shows slight but consistent improvement in solving the swapped word problem. And the method can also be combined with LF-DC-bMMI to achieve 8.2\% relative improvement versus the CE-trained PIT-ASR baseline.

In the eighth to tenth rows, experiments are conducted on 150 hours corpus. The results are similar, and  LF-DC-bMMI criterion shows consistent improvement versus PIT-CE and naive sequence discriminative training criterion. In Section~\ref{Sec:exp-150hrs}, it is shown that sequence discriminative training can be combined with other technologies and achieves further consistent and significant improvement.

The discriminative training criterion helps the system training in two ways.
Firstly, sequence level criterion helps the sequence level speaker tracing problem in PIT modeling. Specifically, linguistic information is encoded in the senone level language modeling in discriminative training. Thus the procedure implicitly integrates linguistic information  in the speaker tracing problem. Secondly, sequence discriminative training improves the speech recognition module. Notably, all the sequence discriminative training procedures are  applied after CE initialization as in~\cite{xiong2016achieving}. With initialization, it also helps Equation~(\ref{equ:bayes-pit}) to reach a better minimum.


\subsection{Combination and Extension to Larger Corpus}
\label{Sec:exp-150hrs}
Table~\ref{tab:exp-combine} summarizes the performance improvement of integrating all the proposed methods. 

\begin{table}[thbp!]
  \caption{\label{tab:exp-combine} {\it  Performance Summary in SWBD 50 Hours Dataset}}
  \centerline{
    \begin{tabular}{ c m{0.33\columnwidth}||c c}
      \hline
      \multicolumn{1}{c}{Neural network } &
      \multicolumn{1}{c||}{Model} &
      \multicolumn{1}{c}{WER} &
      \multicolumn{1}{c}{Rel. (\%) } \\
      \hline \hline
       10$\cdot$0 BLSTM &  PIT-CE& 57.5& 0 \\
        \hline\hline
         \multirow{4}*{{6$\cdot$4 BLSTM}} &progressive joint training & 50.2 & -13 \\
        &\ \ + clean teacher& 38.9&-32.4 \\
        %
        %
        &\ \ + MMI clean teacher&35.8 &-37.7 \\
        &\ \ \ \ + LF-DC-bMMI& 35.2 &  -38.8 \\
      \hline\hline
      \multirow{4}*{{1 LACE + 5$\cdot$4 BLSTM}} &progressive joint training & 47.4 & -17.5 \\
      &  \ \ + clean teacher& 36.0&-37.4\\ 
       &\ \ + MMI clean teacher& 34.6 &-39.8\\ 
       &\ \ \ \ + LF-DC-bMMI& 34.0 & -40.9 \\
      \hline
    \end{tabular}
  }
\end{table}

\begin{figure*}
  \centering
    \includegraphics[width=\linewidth]{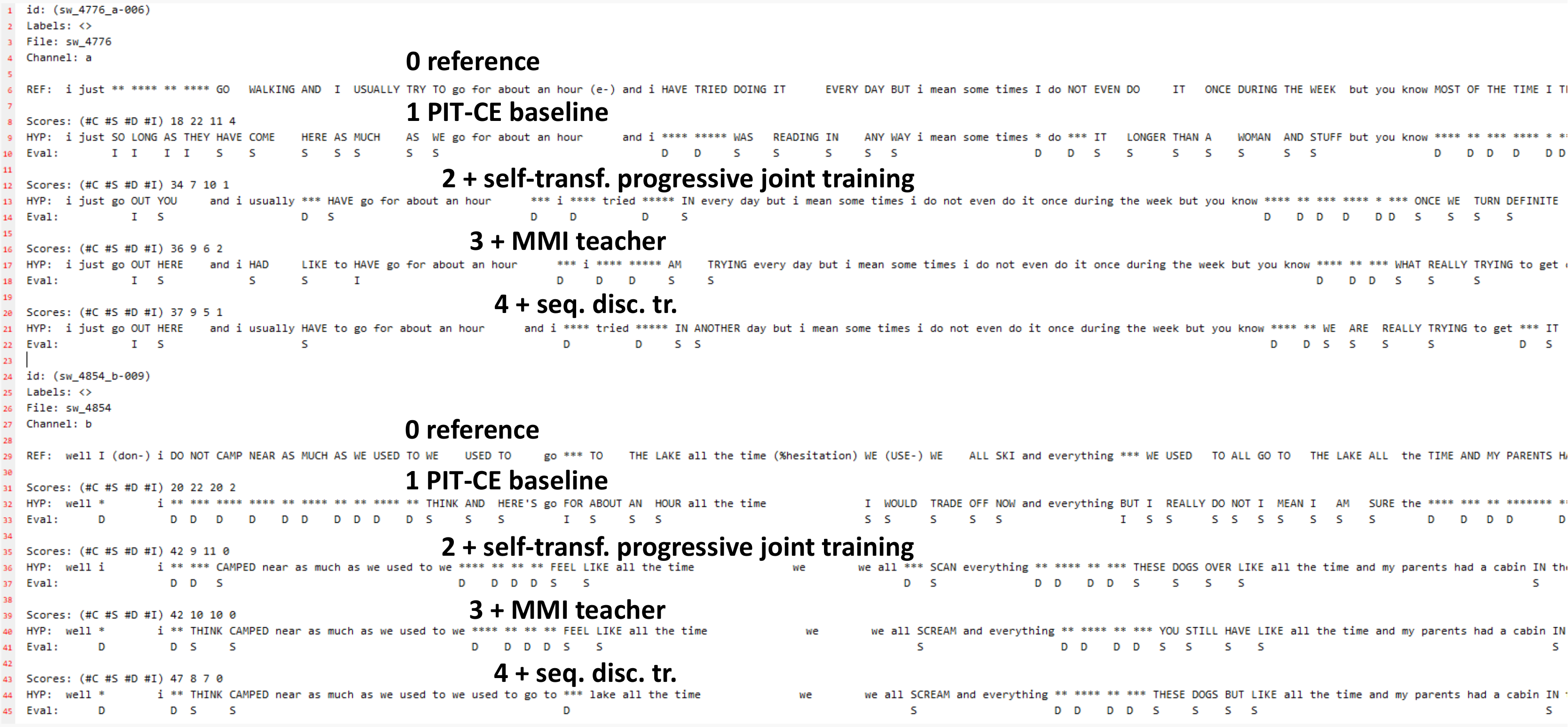}
    \caption{\it  50 hours dataset decoding examples of the proposed methods versus the PIT baseline. The upper part is from stream A and the lower part is from stream B. C, S, D, I refer to correct, substitution, deletion and insertion.}
    \label{fig:example}
\end{figure*}

The PIT-ASR model~\cite{yu2017recognizing}, denoted as PIT-CE, is taken as the baseline of naive joint modeling in the first row. 
The separately optimized system, namely PIT-SS+ASR, is not included here. As shown in Table~\ref{tab:modular}, the performance deteriorates because of feature mismatch.
Instead, the proposed progressive joint training model in the second row can be taken as a stronger separately optimized system with fine-tuning.
The proposed self-transfer learning based joint training model shows further significant improvement in the third and fourth rows. Finally, the multi-output sequence discriminative training is applied and achieves moderate improvement, although the teacher model is already MMI-trained, similar to what has been observed in~\cite{7913606}.
Figure~\ref{fig:example} shows decoding examples of the proposed methods versus the PIT baseline. The baseline contains many errors due to bad model generalization with limited size of dataset. With the proposed methods, errors are significantly reduced. Notably, in this example, the self-transfer learning based progressive joint training mainly reduces errors from similar pronunciations, while sequence discriminative training mainly reduces explicit syntax or linguistic errors, which is  in line with the expectation.

With similar number of parameters but different neural networks, namely {1 LACE + 5$\cdot$4 BLSTM}, the system can be consistently improved in the sixth to ninth rows. We believe the further improvement  comes from the proper modularization of the problem,
which is discussed in Section~\ref{Sec:exp-modular}.



Table~\ref{tab:exp-combine-large} expands the dataset to  150 hours to show the effect of more training data.

\begin{table}[thbp!]
  \caption{\label{tab:exp-combine-large} {\it  Performance Summary in SWBD 150 Hours Dataset}}
  \centerline{
    \begin{tabular}{ c m{0.35\columnwidth}||c c}
      \hline
      \multicolumn{1}{c}{Neural network } &
      \multicolumn{1}{c||}{Model } &
      \multicolumn{1}{c}{WER} &
      \multicolumn{1}{c}{Rel. (\%) } \\
      \hline \hline
       10$\cdot$0 BLSTM&   PIT-CE& 42.2& 0\\
        \hline\hline
       \multirow{3}*{{6$\cdot$4 BLSTM}} &progressive joint training & 41.0& -2.9 \\
        &\ \ + clean teacher& 32.8& -22.3 \\
        &\ \ \ \ + LF-DC-bMMI& 30.8 & -27.0 \\ 
        \hline\hline
        \multirow{3}*{{1 LACE + 5$\cdot$4 BLSTM}} &   progressive joint training & 39.4 & -6.6 \\
        &\ \ + clean teacher& 30.4& -27.9 \\
        &\ \ \ \ + LF-DC-bMMI& 28.0 & -33.6 \\ 
      \hline
    \end{tabular}
  }
\end{table}

The naive joint training baseline in the first row significantly benefits  from more data and shrinks the gap to the proposed progressive joint training model in the second row.
However, it still even significantly worse than the self-transfer learning and sequence discriminative training based joint model trained in 50 hours data in Table~\ref{tab:exp-combine}.
It again shows the disadvantages of large model complexity and insufficient model generalization discussed in Section~\ref{Sec:review-si-ch-rec}. i.e. compared with merely increasing data, the better method to solve the problem is to improve the model generalization.
Besides, the convergence speed of naive joint training model in the larger dataset is even slower, namely 4 times more epochs versus the proposed method. 

Comparing Table~\ref{tab:exp-combine-large} with Table~\ref{tab:exp-combine}, the proposed self-transfer learning based joint training and multi-output sequence discriminative training  show consistent relative
improvement  versus the progressive joint training. Compared with Table~\ref{tab:exp-combine}, sequence discriminative training achieves larger relative improvement on the CE-trained teacher based system.

In both 50 hours and 150 hours corpus, the proposed method  achieves over 30\% relative improvement respectively, versus the PIT-ASR system and the PIT-SS+ASR system.
The improvement comes from better model generalization, training efficiency and the sequence level linguistic knowledge integration.

Although this paper addresses the case of simultaneous speech of two people talking at a relative level of 0dB, we believe it will be straightforward to extend the system to handle more realistic conditions. The case where one speaker is louder than the other has already been observed to be easier for PIT-style models than the 0dB data explored in this paper~\cite{yu2017recognizing}. For more than two speakers, extension of the proposed system should follow the same construction described in~\cite{qian2017single}. 
Finally, we expect robustness to background noise and reverberation to come from standard techniques such as multi-condition training~\cite{seltzer2013investigation}.

\section{Conclusion}\label{Sec:conclu}

In this work, we proposed to divide the single channel overlapped speech recognition problem into three sub-problems: frame-wise interpreting, speaker tracing and speech recognition. Each module is firstly optimized separately with specific designed criteria, which  significantly improves the system generalization and training efficiency. After the initialization, modules are jointly trained with two novel strategies: self-transfer learning and multi-output sequence discriminative training. Specifically, in the joint training stage, the clean speech model fine-tunes its parameters with other modules in overlapped speech to fit its own distribution in the simultaneous clean speech.
And then sequence discriminative training designed for multiple outputs is applied to integrate linguistic and sequence information. The proposed framework achieves  30\% relative improvement over both  a strong jointly trained system, PIT-ASR,  and a separately optimized system, PIT-SS+ASR. 

The proposed framework shows promising perspectives of future improvements, which are: i) Integrating state-of-the-art technologies in each module in the initialization stage, e.g., DPCL~\cite{hershey2016deep}. ii) Applying other sequence level criteria to improve the speaker tracing and speech recognition modules, i.e. connectionist temporal classification (CTC)~\cite{graves2006connectionist}. iii) Explicit integration of language model in the joint modeling, e.g. joint decoding~\cite{weng2015deep} and end-to-end modeling~\cite{hori2017advances}.


%

%
%
%
%
\section*{Acknowledgment}

We thank Chris Basoglu, Frank Seide for their invaluable assistance with CNTK; Mike Seltzer, Takuya Yoshioka, Hakan Erdogan and Andreas Stolcke for many helpful conversations. 
The first author would like to further thank Jerry  and Juncheng Gu for their supports during the internship.

\ifCLASSOPTIONcaptionsoff
  \newpage
\fi



%
%
%

\bibliographystyle{IEEEtran}

  \bibliography{mybib}
%

\vfill

\begin{IEEEbiography}[{\includegraphics[width=1in,height=1.25in,clip,keepaspectratio]{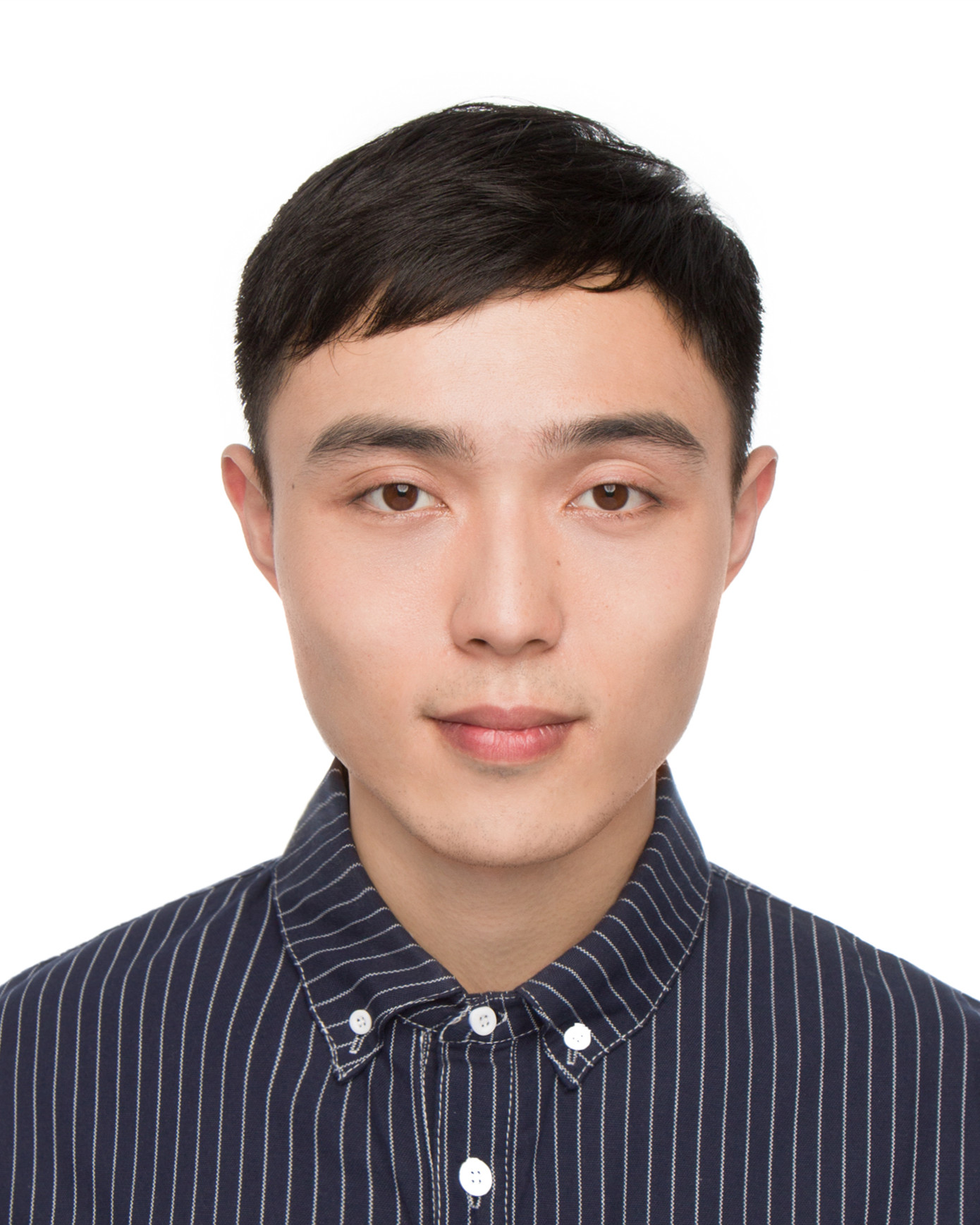}}]{Zhehuai Chen}
received his B.S. degree in
the Department of Electronic and Information
Engineering from Huazhong University of Science
and Technology, China, in 2014.
He is currently a Ph.D. candidate in Shanghai Jiao Tong University working on
speech recognition. His current research interests
include speech recognition, speech synthesis and deep learning.
\end{IEEEbiography}

\vfill

\begin{IEEEbiography}[{\includegraphics[width=1in,height=1.25in,clip,keepaspectratio]{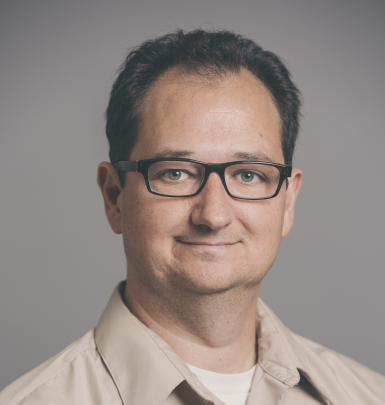}}]{Jasha Droppo}
 (M'03-SM'07) received the B.S.
degree in electrical engineering (with honors) from
Gonzaga University, Spokane, WA, in 1994, and
the M.S. and Ph.D. degrees in electrical engineering from the University of Washington, Seattle, in
1996 and 2000, respectively. At the University of
Washington, he helped to develop and promote a
discrete theory for time-frequency representations of
audio signals, with a focus on speech recognition.
He is best known for his research in robust speech
recognition, including algorithms for speech signal
enhancement, model-based speech feature enhancement, robust speech features, model-based adaptation, and noise tracking. His current interests include
the use of neural networks in acoustic modeling and the application of large
data and general machine learning algorithms to previously hand-authored
speech recognition components.
\end{IEEEbiography}

\vfill

\begin{IEEEbiography}[{\includegraphics[width=1in,height=1.25in,clip,keepaspectratio]{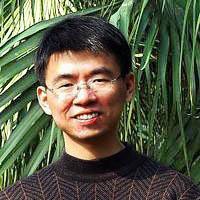}}]{Jinyu Li}
(M'08)  received the bachelor and master degree from University of Science and Technology of
China, in 1997 and 2000, with the highest honor,
and the Ph.D. degree from Georgia Institute of
Technology, Atlanta, in 2008. From 2000 to 2003, he
was a researcher in the Intel China Research Center
and Research Manager in iFlytek Speech, China.
Currently, he is a principal applied scientist and
technical lead in Microsoft Corporation, Redmond,
WA. He leads a team to design and improve speech
modeling algorithms and technologies that ensure
industry state-of-the-art speech recognition accuracy for Microsoft products
such as Cortana and xBox Kinect. His major research interests cover several
topics in speech recognition, including deep learning, noise robustness,
discriminative training, feature extraction, and machine learning methods. He
authored more than 70 refereed publications and around 20 patents. He is
the leading author of the book Robust Automatic Speech Recognition – A
Bridge to Practical Applications-Academic Press, Oct, 2015. Currently, he
serves as the associate editor of IEEE/ACM Transactions on Audio, Speech
and Language Processing.
\end{IEEEbiography}

\vfill

\begin{IEEEbiography}[{\includegraphics[width=1in,height=1.25in,clip,keepaspectratio]{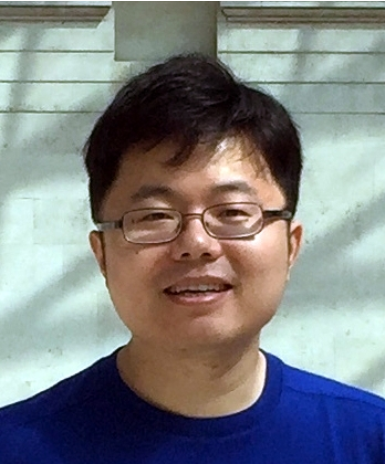}}]{Wayne Xiong}
 (M'07) received the Bachelor degree
in electrical engineering from Tsinghua University,
Beijing, P. R. China, in 2005, and the Master degree
in computer science and engineering from the Chinese University of Hong Kong, Hong Kong, in 2007.
He joined Microsoft Corporation in 2007. He is currently working as a Senior Researcher in the Speech
and Dialog Research Group. His research interests
include speech recognition and next generation of
dialog systems.
\end{IEEEbiography}

\end{document}